\documentclass{article}
\usepackage[symbol]{footmisc}

\usepackage[final]{neurips_2025}  
\usepackage{amsmath}
\usepackage{float} 

\usepackage{algorithm}
\usepackage{algpseudocode} 
\usepackage{comment}

\DeclareMathOperator*{\argmax}{arg\,max} 

\usepackage[utf8]{inputenc} 
\usepackage[T1]{fontenc}    
\usepackage{hyperref}       
\usepackage{url}            
\usepackage{booktabs}       
\usepackage{amsfonts}       
\usepackage{nicefrac}       
\usepackage{microtype}      
\usepackage{xcolor}         
\usepackage{multirow}
\usepackage{graphicx}
\usepackage{fontawesome5} 
\newcommand{\emailsymbol}{\faEnvelope} 

\title{Inference-Time Chain-of-Thought Pruning with Latent Informativeness Signals}

\begin{document}
\author{
  Sophie Li$^{{ \text{ \tiny{\emailsymbol }} 1*}}$
  \And
  Nicholas Huang$^{{ \text{ \tiny{\emailsymbol }} 2*}}$
  \And
  Nayan Saxena$^{3*}$
  \AND
  Nina Luo$^{4}$
  \And
  Vincent Lin$^{5}$
  \And
  Kevin Zhu$^{6}$
  \And
  Sunishchal Dev$^{6}$
}

\maketitle

\begin{abstract}
Large language models (LLMs) improve reasoning accuracy when generating multiple candidate solutions at test time, but standard methods like Best-of-N (BoN) incur high computational cost by fully generating all branches. Self-Truncation Best-of-N (ST-BoN) mitigates this by truncating unpromising paths early, but its reliance on consistency-based heuristics is a limitation as it does not directly evaluate branch quality. We present KL-Adjusted Pruned Path Algorithm (KAPPA), an inference-time method that combines Kullback–Leibler divergence, confidence, and entropy into a principled scoring function to guide progressive pruning. By promoting diversity during exploration and selectively eliminating low-scoring branches, KAPPA maintains accuracy while substantially reducing memory and token usage. Experiments on GSM8K and MATH500 with DeepSeek-R1-Distill-Qwen-1.5B and Qwen2.5-7B-Instruct demonstrate that KAPPA stabilizes performance in smaller models and achieves up to ~60\% reduction in peak memory and ~90\% reduction in total token generation relative to BoN, with minimal impact on accuracy.
\end{abstract}

\section{Introduction}

\footnotetext{\hspace{-2em}$^{*}$Equal contribution. 
$^{1}$Columbia University, $^{2}$University of British Columbia,  $^{3}$ Project Lead, $^{4}$Harvey Mudd College,   $^{5}$University of Florida, $^{6}$Algoverse AI Research \\$\text{\tiny{\emailsymbol}} : $ \texttt{sql2002@columbia.edu; nhuang05@student.ubc.ca} } 
Large Language Models (LLMs) exhibit strong reasoning capabilities thanks to large-scale pretraining \cite{brown2020languagemodelsfewshotlearners}, and chain-of-thought (CoT) prompting further improves reasoning by decomposing problems into intermediate steps \cite{wei2023chainofthoughtpromptingelicitsreasoning}. However, standard autoregressive decoding focuses on locally optimal paths, which can overlook more accurate or efficient reasoning trajectories. Therefore, the inference-time scaling technique \cite{snell2024scalingllmtesttimecompute} is proposed to discover more accurate and efficient paths by sampling multiple candidate reasoning traces. One prominent line of work is Best-of-N (BoN) sampling, where $N$ sequences are sampled independently, scored post-hoc, and the best is chosen: $Y_1,...,Y_N \sim \prod_{t}p_{\theta}(\cdot | x, y_{<t}), \hat{Y} = \argmax_i s(Y_i) $ for some scoring function $s$ \cite{touvron2023llama2openfoundation}. Variants include self-consistency \cite{wang2023selfconsistencyimproveschainthought}, which uses majority voting, and reward-model-based scoring \cite{wang2024interpretablepreferencesmultiobjectivereward}. These methods improve accuracy, but are expensive since every branch must be fully generated \cite{xue-etal-2023-dynamic}. 

The recently proposed Self-Truncation Best-of-N (ST-BoN) method mitigates this inefficiency by truncating all but 1 of the candidate samplings using the early sampling consistency (\cite{wang2025samplingefficienttesttimescalingselfestimating}). ST-BoN performs this truncation once all the samplings reach the earliest point of pairwise difference and have continued to generate for an additional fixed buffer window to allow their divergences to become pronounced. This approach reduces redundant computation and accelerates inference, while retaining the accuracy gains of BoN. However, its consistency-based criterion does not directly assess branch quality, whereas our proposed method applies principled scoring with progressive pruning. 

We introduce KL-Adjusted Pruned Path Algorithm (KAPPA), a novel sampling algorithm whose key features are as follows:
\begin{enumerate}
    \item Exploration vs. efficiency: Diversity is encouraged during the draft phase, while pruning prevents wasted computation on bad branches.
    \item Uncertainty as a self-supervised signal: KL divergence provides a natural, training-free measure of branch informativeness, avoiding the use of external reward models.
    \item Pruning schedule: Progressively pruning branches eliminates unpromising branches earlier, preserving resources for the exploration of more promising branches.
\end{enumerate}

We evaluate KAPPA across multiple reasoning benchmarks (GSM8K, MATH-500) and models (DeepSeek-R1-Distill-Qwen-1.5B, Qwen2.5-7B-Instruct). In our experiments, we demonstrate that KAPPA achieves 72.2\% on MATH500, and reduces token generation by up to 97.3\% compared to standard BoN. Fig. \ref{fig:accuracy} highlights our key results. 

\section{Related Work}

\subsection{Inference-Time Scaling}

Inference-time scaling has been shown to increase reasoning accuracy by allocating more compute at inference-time (\cite{snell2024scalingllmtesttimecompute}). While Chain-of-Thought (CoT) increases accuracy by allocating more compute to a single reasoning trajectory (\cite{wei2023chainofthoughtpromptingelicitsreasoning}), Best-of-N (BoN) sampling searches for more accurate reasoning trajectories by aggregating multiple reasoning trace samples (\cite{snell2024scalingllmtesttimecompute}). A notable paradigm is self-consistency (\cite{wang2023selfconsistencyimproveschainthought}, \cite{jain2024lightweightrerankinglanguagemodel}), which selects the best sample through majority voting. Although this approach does not require any additional training, it significantly increases computational costs as all trace samples must be fully generated before selection. Reward-model-based scoring can further improve performance but adds overhead and risks overoptimization \cite{gao2022scalinglawsrewardmodel}. As such, recent work has explored preemptively ending unpromising reasoning trajectories using LLMs' internal knowledge. Self-Truncation Best-of-N (ST-BoN) does so by using early sampling consistency to truncate all but 1 of the samples once they reach the earliest point of pairwise difference and have continued generating for a fixed buffer window (\cite{wang2025samplingefficienttesttimescalingselfestimating}). Deep Think with Confidence (DeepConf) extends this with adaptive confidence measures, filtering weak traces and performing a confidence-weighted vote over the remainder (\cite{fu2025deepthinkconfidence}). While DeepConf improves both accuracy and efficiency relative to BoN, it still requires many traces to run to completion, making costs significantly higher than greedy decoding.

\subsection{Inference-time Pruning via Token Compression, Steering, or Early Exit}
Recent work leverages mutual information (MI), entropy, and related signals to improve efficiency and compressibility. Mutual Information Preserving Pruning (\cite{westphal2025mutualinformationpreservingneural}) and ACE (\cite{mi2025aceexploringactivationcosine}) operate at the activation level, using metrics like transfer entropy or cosine similarity to retain essential information flow, primarily reducing model size rather than reasoning length.

Complementary approaches analyze token- or step-level reasoning traces directly. Understanding Chain-of-Thought in LLMs Through Information Theory (\cite{ton2025understanding}) quantifies information gain per intermediate step via MI, identifying uninformative or misleading steps. INFORM (Zhou et al., 2024) adaptively determines how many CoT reasoning paths to sample per question based on uncertainty, but does not intervene within paths or explain path quality. Other heuristics include SEAL (\cite{chen2025sealsteerablereasoningcalibration}), a train-free method using latent-space steering vectors, and TokenSkip (\cite{xia2025tokenskipcontrollablechainofthoughtcompression}) and ThinkLess (\cite{li2025thinklesstrainingfreeinferenceefficientmethod}), both aiming to extract answers early.

\subsection{Training-Time}
Model behavior can be adjusted via fine-tuning on curated reasoning trajectories or with stopping signals. Self-Braking Tuning (\cite{zhao2025letllmsbreakfree}) uses a data curation algorithm to select truncation points based on an overthinking metric and fine-tunes models with loss-masked reasoning segments to teach implicit stopping. Preference learning methods, like Chain-of-Preference (\cite{zhang2024chainpreference}), train models to favor shorter, coherent rationales through pairwise comparisons, while THINKPRUNE (\cite{hou2025thinkprune}) employs reinforcement learning (RL) to encourage shorter reasoning chains without sacrificing accuracy. These methods are effective, but model-specific, data-intensive, and offline, making them less general than dynamic, inference-time approaches.

\section{KL-Adjusted Pruned Path Algorithm (KAPPA)}
\begin{algorithm}[!t]
\caption{ KAPPA (compact)}
\begin{algorithmic}[1]
\State \textbf{Input:} model $M_\theta$, prompt $x$, branches $N$, draft cutoff $c$, horizon $\tau$, window $w$, MoM buckets $m$, EMA rate $\alpha$, weights $(w_{\mathrm{KL}},w_C,w_H)$
\State \textbf{Initialize:} alive set $\mathcal{A}_1 \gets \{1,\dots,N\}$; initialize per-branch buffers/statistics
\Statex
\State \textbf{Draft (exploration):}
\For{each branch $i\in\mathcal{A}_1$ \textbf{in parallel}}
  \State sample prefix $y^i_{1:c}$ autoregressively with $p_\theta(\cdot\mid x,y^i_{1:t-1})$
\EndFor
\State generate unconditional logits $q$ from Beginning of Sentence token
\Statex
\State \textbf{Scoring \& Gating (selection over horizon $[c,c+\tau)$):}
\For{$t = c,\dots,c+\tau-1$}
  \For{each alive branch $i\in\mathcal{A}_t$ \textbf{in parallel}}
    \State forward pass $\rightarrow$ get $p_t^i$ (softmax of logits)
    \State compute KL $D_t^i = D_{KL}(p_t^i\|q)$ and information change $\Delta I_t^i \!=\! D_t^i - D_{t-1}^i$
    \State stabilize $\Delta I$ by median-of-means over last $w$ steps (split into $m$ buckets) to get $\hat{\Delta I}_t^i$
    \State apply bias-corrected EMA with rate $\alpha$ to obtain $\mathrm{EMA}_t^i$
    \State compute uncertainty signals: confidence $C_t^i=\max_v p_t^i(v)$ and entropy $H_t^i$
    \State normalize each signal across \emph{alive} branches at time $t$ (z-score, clamp to $[-3,3]$)
    \State form instantaneous score $s_t^i = w_{\mathrm{KL}}\cdot \hat{\mathrm{EMA}}_t^i + w_C\cdot\hat{C}_t^i + w_H\cdot\hat{H}_t^i$
    \State update trajectory score $S_t^i$ by linearly weighting recent $s_{t'}^i$ (weights $\propto t'$)
    \State sample one-step continuation $y_{t+1}^i \sim p_\theta(\cdot\mid x,y_{1:t}^i)$ for the next round
  \EndFor
  \State compute target survivors $R_t = N - \lfloor ((t-c+1)N)/\tau\rfloor$
  \State prune the lowest-scoring branches so $|\mathcal{A}_{t+1}| = R_t$
\EndFor
\Statex
\State \textbf{Continuation (exploitation):}
\State let $i^\star$ be the unique surviving branch; continue decoding it until [EOS]; return $y^{i^\star}_{1:T}$
\end{algorithmic}
\end{algorithm}

\subsection{Explanation of Algorithm}
The algorithm proceeds in three phases. In the Draft Phase, $N$ candidates are generated in parallel until a cutoff timestep $c$,  ensuring sufficient exploration before evaluation. Following the definition in ST-BoN, $c$ is defined as the earliest time that all branches are pairwise inconsistent (\cite{wang2025samplingefficienttesttimescalingselfestimating}). 

During the Scoring phase, each branch $1 \leq i \leq N$ is assessed at every step $t>c$, using three complementary signals: information gain $\Delta I$, confidence $\Delta C$, and entropy $\Delta H$, which are then normalized and combined into a weighted score. These signals are calculated from the intermediate logits of each layer, which are derived from each layer's hidden states by projecting them onto the vocabulary space. The weights of these signals are determined by performing grid search on a subset of the training data. To capture overall branch quality, we compute a trajectory-weighted score $s_{final,t}^i$ that assigns greater weight to more recent tokens, as later steps are more relevant to gauging performance.

During the Gating phase, branches are pruned one at a time on a linear schedule, leaving exactly 1 after $\tau$ steps, which is then generated to completion. The expanded algorithm with complete mathematical details is provided in Appendix \ref{appendix:algorithm}. 

\section{Experiments}

\begin{figure}[!htbp]
    \centering
    \includegraphics[width=1\linewidth]{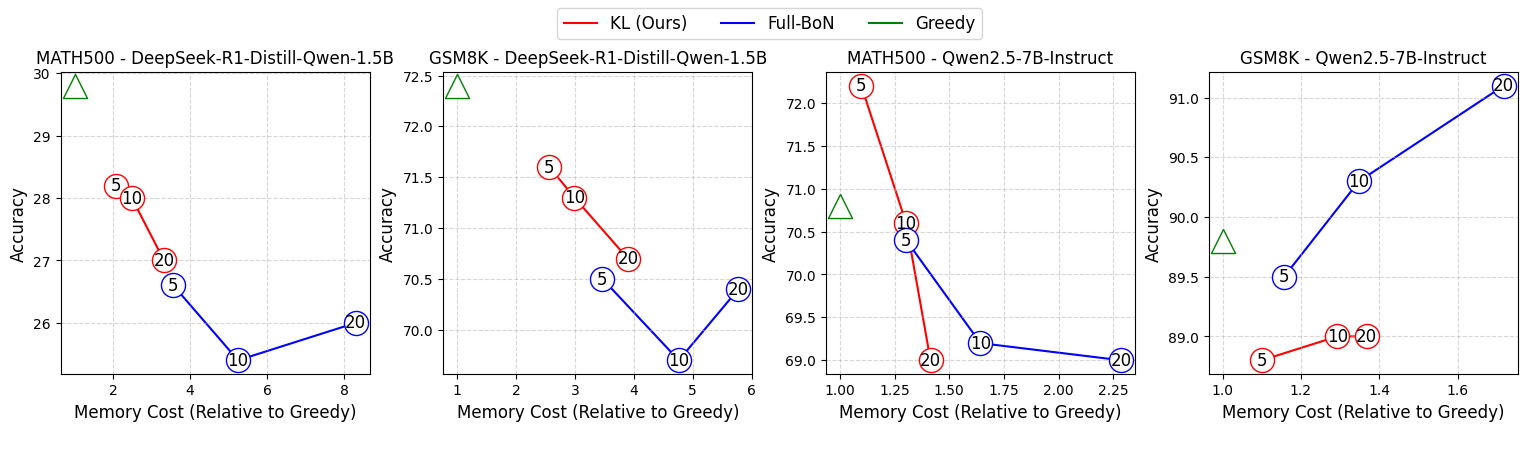}
    \caption{The computational cost and accuracy results in two LLMs across two mathematical and reasoning datasets as labeled. Each point on each polyline represents different sampling sizes $N = 5, 10, 20$ from left to right. }
    \label{fig:accuracy}
\end{figure}
\subsection{Setup}

\textbf{Models.} \quad We evaluate two open-source reasoning LLMs: DeepSeek-R1-Distill-Qwen-1.5B and Qwen-2.5-7B-Instruct. These models are selected for their strong reasoning capabilities given their size. 

\textbf{Benchmarks} \quad We evaluate on two datasets: GSM8K \cite{cobbe2021trainingverifierssolvemath} and MATH500 \cite{hendrycks2021measuringmathematicalproblemsolving, lightman2023letsverifystepstep}. These two datasets cover a variety of difficulty levels of mathematical reasoning problems. 

\textbf{Baselines} \quad We adopt Full-BoN as our primary baseline. Each LLM samples N independent reasoning paths and selects the final answer using the negative perplexity score \cite{kang2025scalablebestofnselectionlarge}.

\textbf{Experimental Settings} \quad We use the sampling strategy that combines top-$k$, top-$p$, and temperature $T$, with $k = 20$, $p = 0.95$, $T$ = 0.7. All baselines are implemented using the HuggingFace Transformers \cite{vaswani2023attentionneed} library's \verb|model.generate()| function. All experiments are run on 80G A100 GPUs, with the number of GPUs varying based on N.

For each model and method, we set the following hyperparameters: Temperature = 0.7, top-$p$ = 0.95, top-$k$ = 20, and max new tokens = 1024. These values are based on the ablation studies conducted for ST-BoN (\cite{wang2025samplingefficienttesttimescalingselfestimating}), which were found to provide the best balance between performance and cost.

Additionally, we set the following hyperparameters for our KL decoding: EMA rate $\alpha = 0.5$, window $w$ = 16, MoM buckets $m$ = 4, and weights $(w_{KL}, w_C, w_H) = (0.7, 0.2, 0.1)$. These hyperparameters were selected through hyperparameter tuning on a subset of the training dataset.

\begin{enumerate}
    \item EMA Rate ($\alpha$): The Exponential Moving Average (EMA) rate determines the weight given to the most recent observations in the moving average calculation. A value of 0.5 provides a balanced approach, allowing the model to be sensitive to recent changes while also considering previously generated tokens.
    \item Window Size ($w$): The window size indicates the number of past observations considered in the moving average. Larger window sizes can smooth out fluctuations but might miss rapid changes, whereas smaller windows capture abrupt shifts but might introduce more noise. A size of 16 was found to show the best performance in our tests.
    \item Median of Means (MoM) Bucket Size ($m$): Median of Means divides data into buckets to compute robust estimates of central tendency. This method is used to decrease sensitivity to outliers, with a bucket size of 4 being shown to provide the best balance between sensitivity and robustness.
    \item Weights for KL, Confidence, and Entropy ($w_{KL}, w_C, w_H$): These weights balance the contribution of different components in the decoding process. A higher weight for KL divergence was found to provide the best performance due to greater emphasis being placed on reducing the difference between model predictions and the actual data distribution, while still having some weight in confidence and entropy.
\end{enumerate}

\textbf{Prompt templates.} \quad For both models, we append the same instruction to every problem prompt: "Please reason step by step, and put your final answer within $\backslash$boxed\{\}." Additionally, we use the official system prompt and put the modified problem in the user message.

In all cases, the final answer is expected to appear inside $\backslash$boxed\{\} and is extracted during post-processing. Decoding terminates only when an end-of-sequence token is produced or the maximum generation length is reached.

\textbf{Evaluation} \quad We assess the balance between performance and peak memory usage during inference. We use the cost of greedy decoding as the baseline, and the memory cost is calculated as follows:

\textit{Memory Cost} $M_{cost}$: Let $M_{peak}$ and $M_{peak}^{greedy}$ denote the peak memory usage during inference. $M_{cost} = \frac{M_{peak}}{M_{peak}^{greedy}}$.

Additionally, we use \textit{Accuracy} as the metric to measure performance. In particular, following \cite{wang2023selfconsistencyimproveschainthought}, we extract answers from LLM responses using an exact match method and compare them with the ground truth. Thus, accuracy is calculated as follows:

\textit{Accuracy A}: Let $N_m$ denote the number of exact matches, and let $N_t$ denote the total number of problems. $A = \frac{N_m}{N_t}$

\subsection{Results}

We present the results of two mathematics datasets in Fig. \ref{fig:accuracy}, and have the following key findings:

\textbf{Reduced memory usage} 
Across all datasets and models, KL consistently lowers peak GPU memory compared to BoN. Memory reductions range from \textasciitilde 4\% to \textasciitilde 60\% relative to BoN. The largest observed difference occurs with using DeepSeek-R1-Distill-Qwen-1.5B on MATH500: KL $N=20$ uses 6495.25MB while BoN $N=20$ uses 16239.977MB. Even at smaller $N$ (e.g., 5 or 10), KL reduces peak memory, proving its efficiency at both low and high sampling scales.

\begin{figure}[!htbp]
    \centering
    \includegraphics[width=1\linewidth]{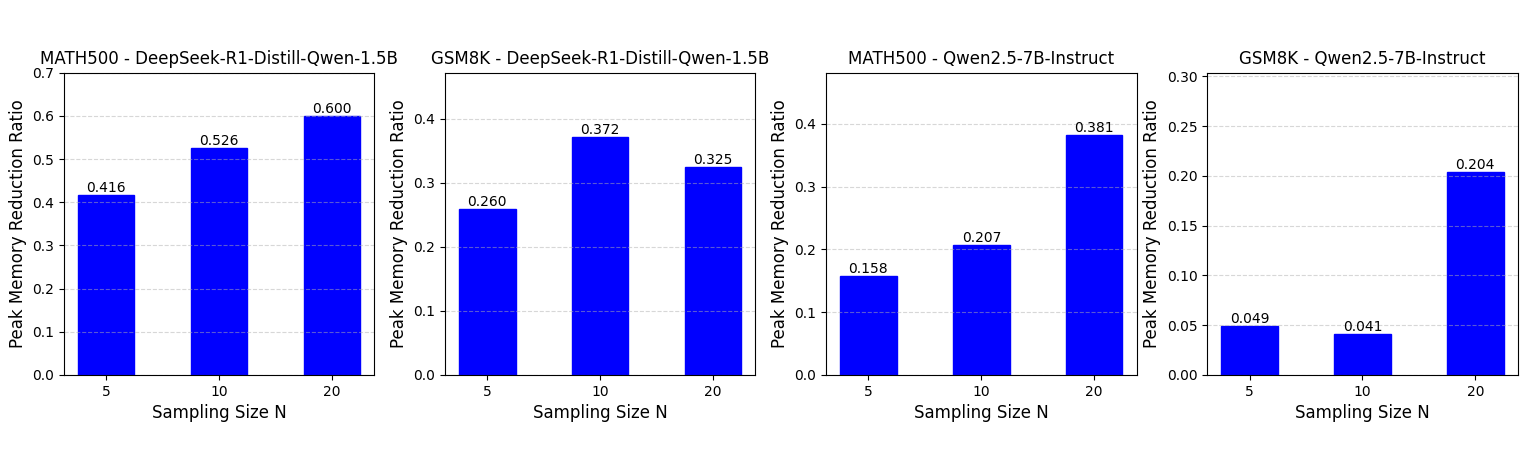}
    \caption{The computed peak memory reduction ratio under different sampling sizes $N$.}
    \label{fig:memory}
\end{figure}

\begin{figure}[!htbp]
    \centering
    \includegraphics[width=1\linewidth]{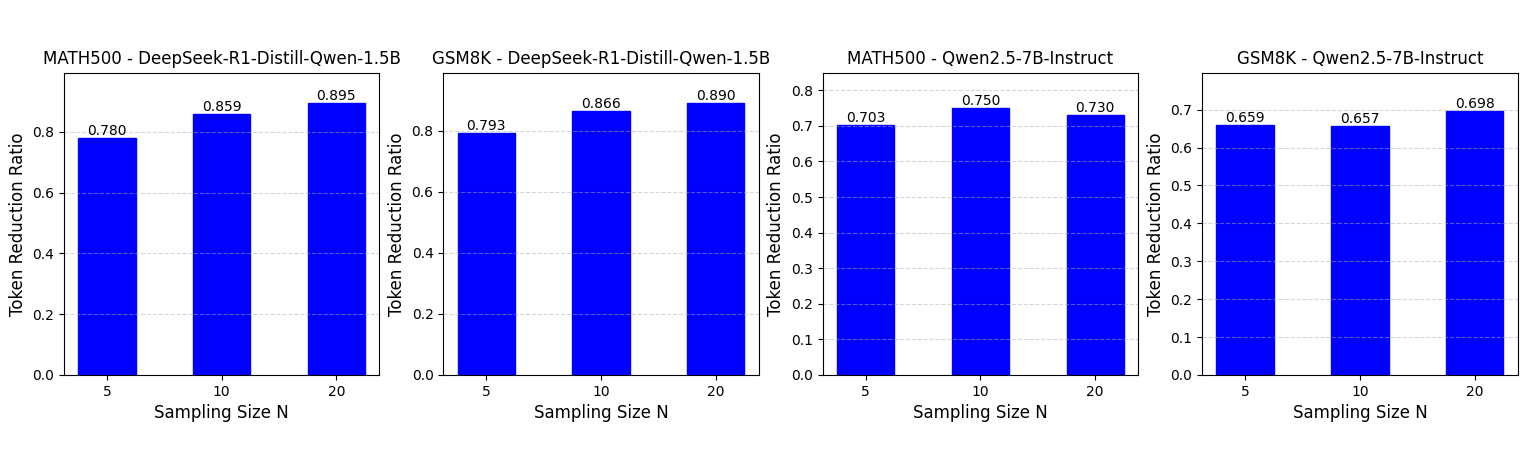}
    \caption{The computed token reduction ratio under different sampling sizes $N$.}
    \label{fig:tokens}
\end{figure}

\textbf{Reduced total tokens} As shown in Fig. \ref{fig:tokens} KL consistently reduces total token generation compared to BoN across all datasets. Token generation reductions range from to \textasciitilde 65\% up to \textasciitilde 90\% relative to BoN. The max difference between the two methods is seen with using DeepSeek-R1-Distill-Qwen-1.5B on MATH500: KL $N=20$ uses $2113.162$ tokens while BoN $N=20$ uses $20053.28$ tokens.

\textbf{Maintained or Improved Accuracy, with Superior Performance on the Small Model} \quad On DeepSeek-R1-Distill-Qwen-1.5B, KL consistently improves accuracy by 1-2\% across both datasets compared to Full-BoN. Additionally, it can achieve these improvements with significantly lower memory cost. These results show that KL is particularly effective at stabilizing performance in smaller models. However, on Qwen2.5-7B-Instruct, we do not see the same consistent improvements as baseline performance strengthens. This indicates that our method's more aggressive branch pruning is effective at cutting out inaccuracies in smaller models, but prunes too aggressively for larger models that have higher quality branches.

In particular, early on in the Scoring and Gating phase for larger models, the lowest scoring branch is often higher quality than the lower scoring branches of smaller models. As such, a linear pruning schedule can eliminate fairly promising branches before they have had the chance to generate towards an accurate answer. This issue can be addressed by extending the draft phase to allow more exploration of each branch before pruning is considered. However, this requires a trade-off between accuracy and cost, which could be addressed by using an adaptive length based on problem complexity. Alternatively, less aggressive pruning schedules, such as a cosine schedule, can be considered to cause less pruning early in the Scoring and Gating phase. To similar effect, the Scoring and Gating phase itself can be extended, perhaps adaptively, to cause less frequent pruning while allowing further exploration of the most promising branches. Finally, further hyperparameter adjustments to reduce sensitivity can be considered, such as lowering the EMA rate $\alpha$, increasing MoM window and bucket sizes, and increasing confidence and entropy weights. Overall, there is a lot of further potential work in tuning the balance between the cost savings of pruning and the accuracy gains of further exploration.

\section{Conclusion}

\paragraph{Future Work \& Limitations} Our current evaluation is limited to specific model scales and datasets, so further experiments with other models and datasets, such as commonsense reasoning datasets, are needed to confirm the generality of our results. Additionally, due to computational restrictions, we were only able to run experiments with max new tokens up to 1024. This mainly affected only the MATH500 experiments on DeepSeek since that model tended to require more than 1024 tokens to reach a final answer on that dataset. Furthermore, the pruning horizon $\tau$ is fixed in this work; dynamically adjusting $\tau$ according to problem complexity could improve scoring quality, which would be reflected in accuracy and token efficiency. Another limitation of our work is that for some models and datasets, the accuracy of KL decreases as N increases. We suspect that this is caused by the over-pruning of promising branches as N increases. To address this, we could experiment with less aggressive pruning schedules, such as a cosine schedule, as well as further hyperparameter tuning to mitigate noisy signals.

\paragraph{Summary of Efficiency Gains} Overall, KL provides a robust tradeoff: notably lower memory usage and token generation without sacrificing accuracy. We illustrate improvements across datasets, models, and sampling counts, highlighting KL’s practical advantage for scaling inference-time decoding efficiently.
\newpage

\bibliographystyle{plainnat}
\bibliography{references}

\begin{thebibliography}{25}
\providecommand{\natexlab}[1]{#1}
\providecommand{\url}[1]{\texttt{#1}}
\expandafter\ifx\csname urlstyle\endcsname\relax
  \providecommand{\doi}[1]{doi: #1}\else
  \providecommand{\doi}{doi: \begingroup \urlstyle{rm}\Url}\fi

\bibitem[Brown et~al.(2020)Brown, Mann, Ryder, Subbiah, Kaplan, Dhariwal, Neelakantan, Shyam, Sastry, Askell, Agarwal, Herbert-Voss, Krueger, Henighan, Child, Ramesh, Ziegler, Wu, Winter, Hesse, Chen, Sigler, Litwin, Gray, Chess, Clark, Berner, McCandlish, Radford, Sutskever, and Amodei]{brown2020languagemodelsfewshotlearners}
Tom~B. Brown, Benjamin Mann, Nick Ryder, Melanie Subbiah, Jared Kaplan, Prafulla Dhariwal, Arvind Neelakantan, Pranav Shyam, Girish Sastry, Amanda Askell, Sandhini Agarwal, Ariel Herbert-Voss, Gretchen Krueger, Tom Henighan, Rewon Child, Aditya Ramesh, Daniel~M. Ziegler, Jeffrey Wu, Clemens Winter, Christopher Hesse, Mark Chen, Eric Sigler, Mateusz Litwin, Scott Gray, Benjamin Chess, Jack Clark, Christopher Berner, Sam McCandlish, Alec Radford, Ilya Sutskever, and Dario Amodei.
\newblock Language models are few-shot learners, 2020.
\newblock URL \url{https://arxiv.org/abs/2005.14165}.

\bibitem[Chen et~al.(2025)Chen, Zhang, Hong, Kundu, and Wang]{chen2025sealsteerablereasoningcalibration}
Runjin Chen, Zhenyu Zhang, Junyuan Hong, Souvik Kundu, and Zhangyang Wang.
\newblock Seal: Steerable reasoning calibration of large language models for free, 2025.
\newblock URL \url{https://arxiv.org/abs/2504.07986}.

\bibitem[Cobbe et~al.(2021)Cobbe, Kosaraju, Bavarian, Chen, Jun, Kaiser, Plappert, Tworek, Hilton, Nakano, Hesse, and Schulman]{cobbe2021trainingverifierssolvemath}
Karl Cobbe, Vineet Kosaraju, Mohammad Bavarian, Mark Chen, Heewoo Jun, Lukasz Kaiser, Matthias Plappert, Jerry Tworek, Jacob Hilton, Reiichiro Nakano, Christopher Hesse, and John Schulman.
\newblock Training verifiers to solve math word problems, 2021.
\newblock URL \url{https://arxiv.org/abs/2110.14168}.

\bibitem[Fu et~al.(2025)Fu, Wang, Tian, and Zhao]{fu2025deepthinkconfidence}
Yichao Fu, Xuewei Wang, Yuandong Tian, and Jiawei Zhao.
\newblock Deep think with confidence, 2025.
\newblock URL \url{https://arxiv.org/abs/2508.15260}.

\bibitem[Gao et~al.(2022)Gao, Schulman, and Hilton]{gao2022scalinglawsrewardmodel}
Leo Gao, John Schulman, and Jacob Hilton.
\newblock Scaling laws for reward model overoptimization, 2022.
\newblock URL \url{https://arxiv.org/abs/2210.10760}.

\bibitem[Hendrycks et~al.(2021)Hendrycks, Burns, Kadavath, Arora, Basart, Tang, Song, and Steinhardt]{hendrycks2021measuringmathematicalproblemsolving}
Dan Hendrycks, Collin Burns, Saurav Kadavath, Akul Arora, Steven Basart, Eric Tang, Dawn Song, and Jacob Steinhardt.
\newblock Measuring mathematical problem solving with the math dataset, 2021.
\newblock URL \url{https://arxiv.org/abs/2103.03874}.

\bibitem[Hou et~al.(2025)Hou, Zhang, Ji, Liu, Qian, Andreas, and Chang]{hou2025thinkprune}
Bairu Hou, Yang Zhang, Jiabao Ji, Yujian Liu, Kaizhi Qian, Jacob Andreas, and Shiyu Chang.
\newblock Thinkprune: Pruning long chain-of-thought of llms via reinforcement learning, 2025.
\newblock URL \url{https://arxiv.org/abs/2504.01296}.

\bibitem[Jain et~al.(2024)Jain, Ma, Deoras, and Xiang]{jain2024lightweightrerankinglanguagemodel}
Siddhartha Jain, Xiaofei Ma, Anoop Deoras, and Bing Xiang.
\newblock Lightweight reranking for language model generations, 2024.
\newblock URL \url{https://arxiv.org/abs/2307.06857}.

\bibitem[Kang et~al.(2025)Kang, Zhao, and Song]{kang2025scalablebestofnselectionlarge}
Zhewei Kang, Xuandong Zhao, and Dawn Song.
\newblock Scalable best-of-n selection for large language models via self-certainty, 2025.
\newblock URL \url{https://arxiv.org/abs/2502.18581}.

\bibitem[Li et~al.(2025)Li, Gao, Li, and Wu]{li2025thinklesstrainingfreeinferenceefficientmethod}
Gengyang Li, Yifeng Gao, Yuming Li, and Yunfang Wu.
\newblock Thinkless: A training-free inference-efficient method for reducing reasoning redundancy, 2025.
\newblock URL \url{https://arxiv.org/abs/2505.15684}.

\bibitem[Lightman et~al.(2023)Lightman, Kosaraju, Burda, Edwards, Baker, Lee, Leike, Schulman, Sutskever, and Cobbe]{lightman2023letsverifystepstep}
Hunter Lightman, Vineet Kosaraju, Yura Burda, Harri Edwards, Bowen Baker, Teddy Lee, Jan Leike, John Schulman, Ilya Sutskever, and Karl Cobbe.
\newblock Let's verify step by step, 2023.
\newblock URL \url{https://arxiv.org/abs/2305.20050}.

\bibitem[Mi et~al.(2025)Mi, Kong, Yuan, and Huang]{mi2025aceexploringactivationcosine}
Zhendong Mi, Zhenglun Kong, Geng Yuan, and Shaoyi Huang.
\newblock Ace: Exploring activation cosine similarity and variance for accurate and calibration-efficient llm pruning, 2025.
\newblock URL \url{https://arxiv.org/abs/2505.21987}.

\bibitem[Snell et~al.(2024)Snell, Lee, Xu, and Kumar]{snell2024scalingllmtesttimecompute}
Charlie Snell, Jaehoon Lee, Kelvin Xu, and Aviral Kumar.
\newblock Scaling llm test-time compute optimally can be more effective than scaling model parameters, 2024.
\newblock URL \url{https://arxiv.org/abs/2408.03314}.

\bibitem[Ton et~al.(2025)Ton, Taufiq, and Liu]{ton2025understanding}
Jean-Francois Ton, Muhammad~Faaiz Taufiq, and Yang Liu.
\newblock Understanding chain-of-thought in {LLM}s through information theory, 2025.
\newblock URL \url{https://openreview.net/forum?id=ouRX6A8RQJ}.

\bibitem[Touvron et~al.(2023)Touvron, Martin, Stone, Albert, Almahairi, Babaei, Bashlykov, Batra, Bhargava, Bhosale, Bikel, Blecher, Ferrer, Chen, Cucurull, Esiobu, Fernandes, Fu, Fu, Fuller, Gao, Goswami, Goyal, Hartshorn, Hosseini, Hou, Inan, Kardas, Kerkez, Khabsa, Kloumann, Korenev, Koura, Lachaux, Lavril, Lee, Liskovich, Lu, Mao, Martinet, Mihaylov, Mishra, Molybog, Nie, Poulton, Reizenstein, Rungta, Saladi, Schelten, Silva, Smith, Subramanian, Tan, Tang, Taylor, Williams, Kuan, Xu, Yan, Zarov, Zhang, Fan, Kambadur, Narang, Rodriguez, Stojnic, Edunov, and Scialom]{touvron2023llama2openfoundation}
Hugo Touvron, Louis Martin, Kevin Stone, Peter Albert, Amjad Almahairi, Yasmine Babaei, Nikolay Bashlykov, Soumya Batra, Prajjwal Bhargava, Shruti Bhosale, Dan Bikel, Lukas Blecher, Cristian~Canton Ferrer, Moya Chen, Guillem Cucurull, David Esiobu, Jude Fernandes, Jeremy Fu, Wenyin Fu, Brian Fuller, Cynthia Gao, Vedanuj Goswami, Naman Goyal, Anthony Hartshorn, Saghar Hosseini, Rui Hou, Hakan Inan, Marcin Kardas, Viktor Kerkez, Madian Khabsa, Isabel Kloumann, Artem Korenev, Punit~Singh Koura, Marie-Anne Lachaux, Thibaut Lavril, Jenya Lee, Diana Liskovich, Yinghai Lu, Yuning Mao, Xavier Martinet, Todor Mihaylov, Pushkar Mishra, Igor Molybog, Yixin Nie, Andrew Poulton, Jeremy Reizenstein, Rashi Rungta, Kalyan Saladi, Alan Schelten, Ruan Silva, Eric~Michael Smith, Ranjan Subramanian, Xiaoqing~Ellen Tan, Binh Tang, Ross Taylor, Adina Williams, Jian~Xiang Kuan, Puxin Xu, Zheng Yan, Iliyan Zarov, Yuchen Zhang, Angela Fan, Melanie Kambadur, Sharan Narang, Aurelien Rodriguez, Robert Stojnic, Sergey Edunov, and Thomas
  Scialom.
\newblock Llama 2: Open foundation and fine-tuned chat models, 2023.
\newblock URL \url{https://arxiv.org/abs/2307.09288}.

\bibitem[Vaswani et~al.(2023)Vaswani, Shazeer, Parmar, Uszkoreit, Jones, Gomez, Kaiser, and Polosukhin]{vaswani2023attentionneed}
Ashish Vaswani, Noam Shazeer, Niki Parmar, Jakob Uszkoreit, Llion Jones, Aidan~N. Gomez, Lukasz Kaiser, and Illia Polosukhin.
\newblock Attention is all you need, 2023.
\newblock URL \url{https://arxiv.org/abs/1706.03762}.

\bibitem[Wang et~al.(2024)Wang, Xiong, Xie, Zhao, and Zhang]{wang2024interpretablepreferencesmultiobjectivereward}
Haoxiang Wang, Wei Xiong, Tengyang Xie, Han Zhao, and Tong Zhang.
\newblock Interpretable preferences via multi-objective reward modeling and mixture-of-experts, 2024.
\newblock URL \url{https://arxiv.org/abs/2406.12845}.

\bibitem[Wang et~al.(2023)Wang, Wei, Schuurmans, Le, Chi, Narang, Chowdhery, and Zhou]{wang2023selfconsistencyimproveschainthought}
Xuezhi Wang, Jason Wei, Dale Schuurmans, Quoc Le, Ed~Chi, Sharan Narang, Aakanksha Chowdhery, and Denny Zhou.
\newblock Self-consistency improves chain of thought reasoning in language models, 2023.
\newblock URL \url{https://arxiv.org/abs/2203.11171}.

\bibitem[Wang et~al.(2025)Wang, Zhang, Huang, Yang, Zhang, Huang, and Wang]{wang2025samplingefficienttesttimescalingselfestimating}
Yiming Wang, Pei Zhang, Siyuan Huang, Baosong Yang, Zhuosheng Zhang, Fei Huang, and Rui Wang.
\newblock Sampling-efficient test-time scaling: Self-estimating the best-of-n sampling in early decoding, 2025.
\newblock URL \url{https://arxiv.org/abs/2503.01422}.

\bibitem[Wei et~al.(2023)Wei, Wang, Schuurmans, Bosma, Ichter, Xia, Chi, Le, and Zhou]{wei2023chainofthoughtpromptingelicitsreasoning}
Jason Wei, Xuezhi Wang, Dale Schuurmans, Maarten Bosma, Brian Ichter, Fei Xia, Ed~Chi, Quoc Le, and Denny Zhou.
\newblock Chain-of-thought prompting elicits reasoning in large language models, 2023.
\newblock URL \url{https://arxiv.org/abs/2201.11903}.

\bibitem[Westphal et~al.(2025)Westphal, Hailes, and Musolesi]{westphal2025mutualinformationpreservingneural}
Charles Westphal, Stephen Hailes, and Mirco Musolesi.
\newblock Mutual information preserving neural network pruning, 2025.
\newblock URL \url{https://arxiv.org/abs/2411.00147}.

\bibitem[Xia et~al.(2025)Xia, Leong, Wang, Li, and Li]{xia2025tokenskipcontrollablechainofthoughtcompression}
Heming Xia, Chak~Tou Leong, Wenjie Wang, Yongqi Li, and Wenjie Li.
\newblock Tokenskip: Controllable chain-of-thought compression in llms, 2025.
\newblock URL \url{https://arxiv.org/abs/2502.12067}.

\bibitem[Xue et~al.(2023)Xue, Liu, Lei, Ren, Yang, Xie, Zhang, Peng, and Lv]{xue-etal-2023-dynamic}
Mingfeng Xue, Dayiheng Liu, Wenqiang Lei, Xingzhang Ren, Baosong Yang, Jun Xie, Yidan Zhang, Dezhong Peng, and Jiancheng Lv.
\newblock Dynamic voting for efficient reasoning in large language models.
\newblock In Houda Bouamor, Juan Pino, and Kalika Bali, editors, \emph{Findings of the Association for Computational Linguistics: EMNLP 2023}, pages 3085--3104, Singapore, December 2023. Association for Computational Linguistics.
\newblock \doi{10.18653/v1/2023.findings-emnlp.203}.
\newblock URL \url{https://aclanthology.org/2023.findings-emnlp.203/}.

\bibitem[Zhang et~al.(2024)Zhang, Du, Pang, Liu, Gao, and Lin]{zhang2024chainpreference}
Xuan Zhang, Chao Du, Tianyu Pang, Qian Liu, Wei Gao, and Min Lin.
\newblock Chain of preference optimization: Improving chain-of-thought reasoning in llms, 2024.
\newblock URL \url{https://arxiv.org/abs/2406.09136}.

\bibitem[Zhao et~al.(2025)Zhao, Yan, Shen, Xu, Zhang, Song, Shao, Lu, Xiao, and Zhuang]{zhao2025letllmsbreakfree}
Haoran Zhao, Yuchen Yan, Yongliang Shen, Haolei Xu, Wenqi Zhang, Kaitao Song, Jian Shao, Weiming Lu, Jun Xiao, and Yueting Zhuang.
\newblock Let llms break free from overthinking via self-braking tuning, 2025.
\newblock URL \url{https://arxiv.org/abs/2505.14604}.

\end{thebibliography}

\newpage 
\section*{Appendix}
 \appendix

\section{Experimental Results}\label{appendix:results}

\scriptsize
\begin{tabular}{ | c | c | c | c | c | c | c | c | c | }
 \hline
 Model & Dataset & Method & N & Accuracy & Final Branch Tokens & Total Tokens & Peak Memory (MB) & Time (s) \\
 \hline
 \multirow{20}{4em}{DeepSeek-R1-Distill-Qwen-1.5B}
 & \multirow{10}{4em}{GSM8K}
  & Greedy & N/A & 0.724 & 434.916 & N/A & 1551.342 & 16.78  \\
 \cline{3-9}
  & & \multirow{3}{4em}{BoN}
   & 5 & 0.705 & 444.309 & 2693.7 & 5378.538 & 13.062 \\
   & & & 10 & 0.697 & 446.114 & 5767.263 & 7389.864 & 7.757 \\
   & & & 20 & 0.704 & 450.953 & 12280.91 & 8962.265 & 11.005 \\
 \cline{3-9}
  & & \multirow{3}{4em}{ST-BoN}
   & 5 & 0.703 & 444.482 & 583.588 & 3667.236 & 10.6 \\
   & & & 10 & 0.692 & 448.221 & 851.474 & 3971.489 & 5.949 \\
   & & & 20 & 0.713 & 442.231 & 1522.451 & 4665.971 & 12.782 \\
 \cline{3-9}
  & & \multirow{3}{4em}{KL}
   & 5 & 0.716 & 447.503 & 557.396 & 3982.059 & 10.994 \\
   & & & 10 & 0.713 & 447.555 & 773.923 & 4637.612 & 9.6 \\
   & & & 20 & 0.707 & 448.284 & 1348.587 & 6046.558 & 9.627 \\
 \cline{2-9}
 & \multirow{10}{4em}{MATH500}
  & Greedy & N/A & 0.298 & 918.536 & N/A & 1955.896 & 37.083 \\
 \cline{3-9}
  & & \multirow{3}{4em}{BoN}
   & 5 & 0.266 & 912.318 & 4895.64 & 6966.972 & 24.605 \\
   & & & 10 & 0.254 & 915.892 & 9936 & 10260.075 & 24.089 \\
   & & & 20 & 0.252 & 917.07 & 20053.28 & 16239.977 & 14.49 \\
 \cline{3-9}
  & & \multirow{3}{4em}{ST-BoN}
   & 5 & 0.254 & 923.134 & 1098.158 & 3711.102 & 10.835 \\
   & & & 10 & 0.26 & 911.796 & 1428.306 & 4080.715 & 22.419 \\
   & & & 20 & 0.256 & 913.686 & 2227.878 & 4877.766 & 11.776 \\
 \cline{3-9}
  & & \multirow{3}{4em}{KL}
   & 5 & 0.282 & 943.006 & 1078.286 & 4065.271 & 11.191 \\
   & & & 10 & 0.28 & 965.054 & 1398.098 & 4861.388 & 22.87 \\
   & & & 20 & 0.27 & 979.356 & 2113.162 & 6495.25 & 24.899 \\
 \hline
\end{tabular}

\scriptsize
\begin{tabular}{ | c | c | c | c | c | c | c | c | c | }
 \hline
 Model & Dataset & Method & N & Accuracy & Final Branch Tokens & Total Tokens & Peak Memory (MB) & Time (s) \\
 \hline
 \multirow{20}{4em}{Qwen2.5-7B-Instruct}
 & \multirow{10}{4em}{GSM8K}
  & Greedy & N/A & 0.898 & 297.187 & N/A & 6633.562 & 18.471 \\
 \cline{3-9}
  & & \multirow{3}{4em}{BoN}
   & 5 & 0.895 & 298.094 & 1531.536 & 7674.957 & 15.661 \\
   & & & 10 & 0.903 & 297.193 & 3064.939 & 8945.062 & 16.307 \\
   & & & 20 & 0.911 & 299.784 & 7816.513 & 11396.753 & 14.429 \\
 \cline{3-9}
  & & \multirow{3}{4em}{ST-BoN}
   & 5 & 0.9 & 299.471 & 547.146 & 6772.066 & 8.878 \\
   & & & 10 & 0.901 & 298.017 & 1040.697 & 7266.462 & 9.097 \\
   & & & 20 & 0.889 & 299.205 & 2232.667 & 8343.815 & 12.834 \\
 \cline{3-9}
  & & \multirow{3}{4em}{KL}
   & 5 & 0.888 & 302.782 & 522.306 & 7299.262 & 6.345 \\
   & & & 10 & 0.89 & 304.26 & 1052.52 & 8577.136 & 11.236 \\
   & & & 20 & 0.89 & 303.227 & 2362.459 & 9075.774 & 13.495 \\
 \cline{2-9}
 & \multirow{10}{4em}{MATH500}
  & Greedy & N/A & 0.708 & 570.904 & N/A & 6811.377 & 27.544 \\
 \cline{3-9}
  & & \multirow{3}{4em}{BoN}
   & 5 & 0.704 & 576.008 & 2917.364 & 8874.21 & 22.97 \\
   & & & 10 & 0.692 & 575.376 & 5808.07 & 11187.881 & 23.431 \\
   & & & 20 & 0.69 & 568.028 & 14417.24 & 15571.764 & 28.485 \\
 \cline{3-9}
  & & \multirow{3}{4em}{ST-BoN}
   & 5 & 0.69 & 573.542 & 880.974 & 6835.073  & 24.815 \\
   & & & 10 & 0.708 & 567.824 & 1409.036 & 7372.32 & 25.3 \\
   & & & 20 & 0.706 & 574.536 & 2592.588 & 8487.64 & 25.557 \\
 \cline{3-9}
  & & \multirow{3}{4em}{KL}
   & 5 & 0.722 & 574.39 & 866.526 & 7476.336 & 17.9 \\
   & & & 10 & 0.706 & 576.372 & 1452.918 & 8872.985 & 20.176 \\
   & & & 20 & 0.69 & 586.89 & 2887.866 & 9633.593 & 24.05 \\
 \hline
\end{tabular}
\normalsize

\newpage

\section{Expanded Algorithm}\label{appendix:algorithm}

 \begin{algorithm}[H]
\caption{ KAPPA (expanded)}
\label{alg:rigel}
\begin{algorithmic}[1]
\Require model $M_\theta$, prompt $x$, number of branches $N$, draft cutoff $c$, pruning horizon $\tau$, window size $w$, MoM buckets $m$, EMA rate $\alpha\!\in\!(0,1)$, weights $(w_{\mathrm{KL}}, w_C, w_H)$
\Ensure sequence $\hat{y}$ decoded by a single surviving branch
\State \textbf{Notation:} vocabulary $V$; alive set $\mathcal{A}_t \subseteq \{1,\ldots,N\}$; per-branch distribution $p_t^i \in \Delta^{|V|-1}$; reference distribution $q \in \Delta^{|V|-1}$ (e.g., teacher, ensemble, or pooled baseline)

\Statex\hfill\textit{// Phase I: Draft (Exploration)}
\State Initialize $\mathcal{A}_1 \gets \{1,\ldots,N\}$; for each $i\in\mathcal{A}_1$ set $y^i_{1:0} \gets \emptyset$
\For{$t = 1$ to $c$}
    \For{each $i \in \mathcal{A}_t$ \textbf{in parallel}}
        \State Sample $y_t^i \sim p_\theta(\cdot \mid x, y_{1:t-1}^i)$ and set $y_{1:t}^i \gets (y_{1:t-1}^i, y_t^i)$
    \EndFor
    \State $\mathcal{A}_{t+1} \gets \mathcal{A}_t$
\EndFor
\State generate unconditional logits $q$ from Beginning of Sentence token

\Statex\hfill\textit{// Phase II: Scoring \& Gating (Selection over a horizon)}
\State Initialize per-branch statistics for $t=c$ (buffers for $\{\Delta I_j^i\}$, MoM blocks, EMA, etc.)
\For{$t = c$ \textbf{to} $c+\tau-1$}
    \For{each $i \in \mathcal{A}_t$ \textbf{in parallel}}
        \State \textbf{Forward pass:} obtain logits from $M_\theta$ and compute $p_t^i = \mathrm{softmax}(\mathrm{logits}_t^i)$
        \State \textbf{Information change:} \(
        D_t^i \gets D_{\mathrm{KL}}(p_t^i \,\|\, q), \quad
        \Delta I_t^i \gets D_t^i - D_{t-1}^i
        \) \hfill\(\triangleright\) $D_{c-1}^i \equiv 0$ for initialization
        \State \textbf{Robustification (Median-of-Means):} partition $\{\Delta I_j^i\}_{j=t-w+1}^{t}$ into $m$ equal-size buckets $\{B_k\}_{k=1}^m$;\\
        \hskip\algorithmicindent \(
        \hat{\Delta I}_t^i \gets \mathrm{median}\Big\{ \frac{1}{|B_k|}\!\sum_{j \in B_k}\!\Delta I_j^i \Big\}_{k=1}^{m}
        \)
        \State \textbf{Temporal smoothing (bias-corrected EMA):}
        \[
        \mathrm{EMA}_t^i \gets \frac{\alpha \hat{\Delta I}_t^i + (1-\alpha)\,\mathrm{EMA}_{t-1}^i}{1-(1-\alpha)^{t-c+1}}, \quad
        \mathrm{EMA}_{c-1}^i \gets 0
        \]
        \State \textbf{Uncertainty signals:} \(
        C_t^i \gets \max_{v \in V} p_t^i(v), \quad
        H_t^i \gets - \sum_{v \in V} p_t^i(v)\log (p_t^i(v) + \varepsilon)
        \)
        \State \textbf{Across-branch normalization (per time $t$):} for each $z \in \{\mathrm{EMA}, C, H\}$,
        \[
        \tilde{z}_t^i \gets \frac{z_t^i - \mu_t(z)}{\sigma_t(z) + \varepsilon}, \quad
        \hat{z}_t^i \gets \mathrm{clip}(\tilde{z}_t^i, -3, 3)
        \]
        \State \textbf{Instantaneous aggregate score:} \(
        s_t^i \gets w_{\mathrm{KL}}\hat{\mathrm{EMA}}_t^i + w_C \hat{C}_t^i + w_H \hat{H}_t^i
        \)
        \State \textbf{Trajectory-weighted score:} \(
        S_t^i \gets \sum_{t'=c}^{t} \omega_{t',t}\, s_{t'}^i, \;\;
        \omega_{t',t} \propto t', \;\; \sum_{t'=c}^{t}\omega_{t',t}=1
        \) \hfill\(\triangleright\) e.g., $\omega_{t',t}=\frac{t'}{\sum_{u=c}^{t}u}$
        \State \textbf{One-step continuation (for next scoring step):} sample $y_{t+1}^i \sim p_\theta(\cdot \mid x, y_{1:t}^i)$
    \EndFor
    \State \textbf{Linear prune schedule:} target survivors
    \(
    R_t \gets N - \Big\lfloor \frac{(t-c+1)N}{\tau} \Big\rfloor
    \)
    \State Remove the $|\mathcal{A}_t|-R_t$ branches with smallest $S_t^i$; set $\mathcal{A}_{t+1}$ to the survivors
\EndFor

\Statex\hfill\textit{// Phase III: Continuation (Exploitation)}
\State Let $i^\star \in \mathcal{A}_{c+\tau}$ be the unique surviving branch (break ties by larger $S_{c+\tau-1}^i$ then lexicographic)
\While{$y^{{i^\star}}_{t} \neq \text{[EOS]}$}
    \State Greedy or sampled decoding with $M_\theta$ conditioned on $(x, y^{i^\star}_{1:t-1})$; append $y_t^{i^\star}$
\EndWhile
\State \textbf{Return} $\hat{y} \gets y^{i^\star}_{1:T}$
\end{algorithmic}
\end{algorithm}
\end{document}